\documentclass[10pt,journal]{IEEEtran}
\ifCLASSOPTIONcompsoc
  \usepackage[nocompress]{cite}
\else
  \usepackage{cite}
\fi
\ifCLASSINFOpdf
\else
\fi

\usepackage{hyperref}
\usepackage{graphicx}
\usepackage{caption}
\usepackage[compact]{titlesec}
\usepackage{subfigure}
\usepackage{graphicx}

\usepackage{amsmath,amsthm}
\usepackage{amssymb,wasysym}
\usepackage{mathrsfs}
\usepackage{cite}
\usepackage{color}
\usepackage{url}
\usepackage{booktabs}
\usepackage{makecell}
\usepackage{multirow}
\usepackage{wrapfig}
\usepackage{verbatim}
\usepackage{dsfont}
\usepackage{amsmath}
\usepackage{algorithm}
 \usepackage{algorithmic}
\usepackage{xurl}

\renewcommand{\algorithmicrequire}{\textbf{Input:}}  % Use Input in the format of Algorithm
\renewcommand{\algorithmicensure}{\textbf{Output:}} % Use Output in the format of Algorithm
\allowdisplaybreaks[3]

\ifodd 1
\else

\fi

\begin{document}

\title{SplitLoRA: A Split Parameter-Efficient Fine-Tuning Framework for Large Language Models}

\author{Zheng Lin, Xuanjie Hu, Yuxin Zhang, Zhe Chen,~\IEEEmembership{Member,~IEEE}, Zihan Fang, Xianhao Chen,~\IEEEmembership{Member,~IEEE}, Ang Li,~\IEEEmembership{Member,~IEEE}, Praneeth Vepakomma,~\IEEEmembership{Member,~IEEE}, and Yue Gao,~\IEEEmembership{Fellow,~IEEE}
\thanks{Z. Lin, X. Hu, Y. Zhang, Z. Chen, Z. Fang and Y. Gao are with the School of Computer Science, Fudan University, Shanghai 200438, China (e-mail: zlin20@fudan.edu.cn; xjhu23@m.fudan.edu.cn; yuxinzhang22@m.fudan.edu.cn; zhechen@fudan.edu.cn; zhfang19@fudan.edu.cn; gao.yue@fudan.edu.cn). Z. Lin is also with the Department of Electrical and Electronic Engineering, University of Hong Kong, Pok Fu Lam, Hong Kong, China. }
\thanks{X. Chen is with the Department of Electrical and Electronic Engineering,
University of Hong Kong, Pok Fu Lam, Hong Kong, China (e-mail: xchen@eee.hku.hk).}
\thanks{A. Li is with the Department of Electrical and Computer Engineering,
University of Maryland, College Park, MD-20742, USA (e-mail: angliece@umd.edu).}
\thanks{P. Vepakomma is with Mohamed bin Zayed University of Artificial Intelligence, Abu Dhabi, United Arab Emirates, and the Massachusetts Institute of Technology, Cambridge, MA 02139 USA (e-mail: vepakom@mit.edu).}
\thanks{\textit{(Corresponding author: Yue Gao)}}
}

\markboth{Journal of \LaTeX\ Class Files,~Vol.~14, No.~8, August~2015}%
{Shell \MakeLowercase{\textit{et al.}}: Bare Advanced Demo of IEEEtran.cls for IEEE Computer Society Journals}


\IEEEtitleabstractindextext{
\begin{abstract}
The scalability of large language models (LLMs) in handling high-complexity models and large-scale datasets has led to tremendous successes in pivotal domains. While there is an urgent need to acquire more training data for LLMs, a concerning reality is the depletion of high-quality public datasets within a few years. In view of this, the federated learning (FL) LLM fine-tuning paradigm recently has been proposed to facilitate collaborative LLM fine-tuning on distributed private data, where multiple data owners collaboratively fine-tune a shared LLM without sharing raw data. However, the staggering model size of LLMs imposes heavy computing and communication burdens on clients, posing significant barriers to the democratization of the FL LLM fine-tuning paradigm. To address this issue, split learning (SL) has emerged as a promising solution by offloading the primary training workload to a server via model partitioning while exchanging activation/activation's gradients with smaller data sizes rather than the entire LLM. Unfortunately, research on the SL LLM fine-tuning paradigm is still in its nascent stage. To fill this gap, in this paper, we propose the first SL LLM fine-tuning framework, named SplitLoRA. SplitLoRA is built on the split federated learning (SFL) framework, amalgamating the advantages of parallel training from FL and model splitting from SL and thus greatly enhancing the training efficiency. It is worth noting that SplitLoRA is the inaugural open-source benchmark for SL LLM fine-tuning, providing a foundation for research efforts dedicated to advancing SL LLM fine-tuning. Extensive simulations validate that SplitLoRA achieves target accuracy in significantly less time than state-of-the-art LLM fine-tuning frameworks, demonstrating the superior training performance of SplitLoRA. The project page is available at \href{https://fdu-inc.github.io/splitlora/}{https://fdu-inc.github.io/splitlora/}.
\end{abstract}

}

\maketitle

% is not selected <OR> if conference mode is selected - because compsoc
% conference papers position the abstract like regular (non-compsoc)
\IEEEdisplaynontitleabstractindextext

% \fi
\IEEEpeerreviewmaketitle

% 1.5 pages
\section{Introduction}\label{Intro}

In recent years, LLMs~\cite{openai2023gpt4,llama2,palm} have achieved tremendous successes across a broad spectrum of pivotal domains such as smart healthcare~\cite{cardenas2024autohealth,lin2023pushing}, computer vision~\cite{berrios2023towards,qiu2024ifvit,wang2024visionllm}, intelligent transportation~\cite{hu2024agentscodriver,zhou2023vision,tian2023vistagpt} due to their exceptional capabilities in handling high-complexity models and large-scale datasets~\cite{openai2023gpt4,llama2,palm,qu2024trimcaching1}. However, a critical concern accompanying the surge of LLMs has emerged: it is estimated that high-quality public datasets will be depleted before 2026~\cite{villalobos2022will}. The increasing trend of researchers preferring to train data-hungry LLMs by combining existing datasets~\cite{wang2023far} or utilizing model-generated datasets~\cite{xu2023wizardlm}, rather than collecting or generating new dataset, also reflects the current scarcity of public available data. This suggests that the development of current LLMs may encounter significant bottlenecks shortly, as the well-established scaling laws indicate that larger datasets usually lead to better performance~\cite{kaplan2020scaling}.

With the rapid proliferation of Internet of Things (IoT) devices and advancements in sensor technology, connected IoT devices are capable of collecting massive data.  However, these high-quality data distributed across multiple parties cannot be shared publicly due to issues such as user privacy concerns (e.g., medical~\cite{llm_medicine} and financial~\cite{wu2023bloomberggpt} data) or physical limitations (e.g., lack of network connectivity). Some leading AI technology companies can gather large volumes of private data to train data-hungry LLMs, exemplified by Google's Med-PaLM~\cite{singhal2023towards}, a medical LLM capable of providing expert-level medical guidance and diagnosis. Nevertheless, not every party possesses adequate private data to independently train a high-performing and data-hungry LLM. Considering the increasing scarcity of public data and the difficulties in accessing user-private data, devising collaborative training paradigms for decentralized private data without data sharing is paramount to driving the advancement of modern LLMs.

The federated learning (FL) LLM training/fine-tuning paradigm~\cite{ye2024openfedllm,cai2023efficient,fang2024automated} has recently been proposed to facilitate collaborative LLM training on distributed private data, where multiple data owners collaboratively training a shared LLM without sharing raw data. In FL LLM paradigm, data owners train local LLMs on their respective local data and then send the LLM parameters rather than raw data to a parameter server for model update~\cite{lin2023fedsn,konevcny2016federated,zhang2024fedac}. However, the staggering growth in LLM model sizes imposes heavy computing and communication burdens, posing significant barriers to the democratization of the FL LLM paradigm. To address this issue, split learning (SL)~\cite{vepakomma2018split} has emerged as a promising distributed ML framework capable of overcoming the weakness of FL by offloading the primary training workload to a server via model partitioning while exchanging activation/activation's gradients with smaller data sizes rather than the entire LLM~\cite{lin2023efficient,lyu2023optimal,lin2023split}. Unfortunately, research on the SL LLM fine-tuning paradigm is still unexplored.

To fill this gap, in this paper, we propose a concise and research-friendly SL LLM fine-tuning framework, named SplitLoRA. SplitLoRA is built on the split federated learning (SFL)~\cite{thapa2022splitfed} framework that amalgamates the advantages of parallel training from FL and model splitting from SL and is developed on the well-known parameter-efficient fine-tuning (PEFT) technique Low-rank adaptation (LoRA)~\cite{hu2021lora,sheng2023s}. Specifically, we start from pre-trained LLMs on a large-scale dataset and then train/fine-tune the LLMs to adapt the downstream tasks via SFL, which consists of several stages: client-side forward propagation (FP), activations transmissions, server-side FP and back-propagation (BP), activations' gradients transmissions, client-side BP, and client-side model aggregation. For experimental configurations, we 1) implement SplitLoRA framework based on GPT-2~\cite{radford2019language} on the E2E dataset~\cite{novikova2017e2e} and evaluate its performance across diverse performance metrics (e.g., BLEU, METEOR, and NIST); 2) adopt the SFL framework, amalgamating the advantages of parallel training from FL and model splitting from SL and thus greatly enhancing the training efficiency. 3) employ the PEFT technique, LoRA, enabling training to be executed on a single consumer-grade GPU (such as the NVIDIA 3090). Besides, we investigate the potential of SplitLoRA for practical deployments by comparing it with state-of-the-art LLM fine-tuning paradigm in terms of converged accuracy, convergence rate, and resource efficiency. It is worth noting that SplitLoRA is the inaugural open-source benchmark for SL LLM fine-tuning, providing a foundation for research efforts dedicated to advancing SL LLM fine-tuning. 

In the near future, we anticipate that others will build upon our SplitLoRA framework for further explorations. There are several reasons for this: (1) Some new challenges and directions are emerging in the SL LLM fine-tuning, such as determining optimal model splitting for LLMs and extension of SL LLM paradigm to accommodate heterogeneous computing resources across clients. (2) Tailoring the SL LLM training framework specifically to different application scenarios, e.g., vehicular networks and satellite networks, etc., to achieve higher training efficiency. (3) In this era of LLMs, we advocate future work in the SL LLM domain to modify our framework and assess the performance of their algorithms.

Our contributions are as follows:

\begin{itemize}
\item We explore the pipeline for fine-tuning contemporary LLMs on decentralized private data resources via SFL, pointing out a promising development direction for LLMs.
\item We propose a concise and research-friendly SL LLM fine-tuning framework named SplitLoRA. It is worth noting that SplitLoRA is the inaugural open-source benchmark for SL LLM fine-tuning, providing a foundation for research efforts dedicated to advancing SL LLM fine-tuning.
\item We conduct extensive experiments to compare SplitLoRA with the conventional centralized and federated LLM fine-tuning paradigms in terms of converged accuracy, convergence rate, and resource efficiency, demonstrating that SplitLoRA is more communication- and computation-efficient.
\end{itemize}

The remainder of this paper is organized as follows. Section~\ref{Rel_Work} introduces related work. Section~\ref{spl_llm} elaborates on system model and SplitLoRA framework.
Section~\ref{experiment} provides the simulation results.
Section~\ref{future_dir} discusses the future research direction.
Finally, concluding remarks are presented in Section~\ref{conclu}.

\section{Related Work}\label{Rel_Work}
\subsection{Large Language Models}

Driven by the maturation of deep learning algorithms, increased computing capabilities, and the availability of large-scale datasets, LLMs have made significant strides across industries. Major players in the AI community~\cite{openai2023gpt4,llama2,palm,dale2021gpt,sanderson2023gpt,devlin2018bert}, including OpenAI, Google, Microsoft, and Facebook, are dedicated to developing their own LLMs. For instance, OpenAI's highly acclaimed chat LLM, GPT series~\cite{openai2023gpt4,dale2021gpt,sanderson2023gpt}, and Google's BERT~\cite{devlin2018bert} have demonstrated superior performance in a wide spectrum of Natural Language Processing (NLP) tasks like language translation, text generation, question answering~\cite{jiang2023fdapt}, and sentiment analysis~\cite{englhardt2023classification, nan2021deep}. Moreover, LLM has extended beyond its original NLP domain to shine in pivotal domains including healthcare~\cite{singhal2023towards,cardenas2024autohealth}, autonomous driving~\cite{wang2023drivemlm,cui2024survey}, and robotic control~\cite{mai2023llm,kannan2023smart}. For example, in healthcare, Med-PaLM~\cite{singhal2023towards} is devised for medical image analysis, clinical document processing, and patient diagnosis, facilitating accurate diagnoses and treatment decisions of healthcare professionals. In the realm of autonomous driving, DriveMLM~\cite{wang2023drivemlm} bridges the gap between language decisions and vehicle control commands, enabling closed-loop autonomous driving in realistic simulators. As the scale of LLM models substantially increases, they exhibit exceptional generalization capabilities—a phenomenon known as ``emergence"~\cite{lin2023pushing}. One of the most representative examples is GPT-4~\cite{sanderson2023gpt}, due to its staggering model size, can successfully perform arithmetic operations such as numerical multiplication without specific targeted training. The outstanding capabilities of LLMs enable them to be directly applied or easily adapted (e.g., through fine-tuning or instruction adjustment) to various downstream or novel tasks, thereby unleashing unprecedented potential in applications such as chatbots, healthcare assistants, and intelligent transportation~\cite{wang2023drivemlm,lin2023pushing,openai2023gpt4,hu2023adaptive,lin2022channel,hu2024collaborative}.

\subsection{Split Learning}
% \rev{
% Considering the increasing scarcity of public data and the difficulties in accessing user-private data, developing collaborative training paradigms for decentralized private data without data sharing is paramount to driving the advancement of modern LLMs.
% }

Split learning (SL)~\cite{vepakomma2018split} has emerged as a promising distributed ML framework, capable of overcoming the weaknesses of FL~\cite{mcmahan2017communication,konevcny2016federated}. SL offloads the primary training workload to the server via model partitioning~\cite{lin2023split,thapa2022splitfed,lin2024adaptsfl}, while exchanging activation/activation's gradients with smaller data sizes rather than the entire model, thereby significantly reducing client-side computing costs and communication overhead during model training~\cite{lyu2023optimal,lin2023split}. However, the original SL, entitled vanilla SL~\cite{vepakomma2018split}, employs a sequential training mechanism from one device to another, substantially reducing training efficiency and resulting in excessive training latency.

To address this issue, split federated learning (SFL)~\cite{thapa2022splitfed} has been proposed, which merges the advantages of parallel training from FL and model splitting from SL.  Apart from model splitting, SFL features periodic server-side and client-side sub-model aggregation to achieve model synchronization after multiple training rounds, aligning with the design principle of FL~\cite{lin2023split,lin2023efficient}. Due to its salient advantages, SFL has garnered significant attention from academia and industry in recent years. In academia, the predominant research efforts focus on the framework design of SFL to enhance model training efficiency, including model splitting~\cite{lin2024adaptsfl,wu2023split}, model aggregation~\cite{lin2024adaptsfl}, client selection~\cite{liu2022wireless}, and device clustering~\cite{wu2023split,cheng2023cheese}. In the industry, Huawei deems SFL as a pivotal learning framework for 6G edge intelligence due to its flexibility and inherent advantages~\cite{huawei2019}.

The effectiveness of SFL methods has been validated in image classification and small-scale models (e.g., ResNet-18~\cite{he2016deep} and VGG-16~\cite{simonyan2014very}). However, the performance of SFL frameworks in the current LLM training paradigm remains unclear. Therefore, in this study, we are the first to investigate the performance of SFL frameworks in LLM training, providing novel insights into SL LLM training and establishing a foundation for future research.

\subsection{Parameter-Efficient Fine-Tuning}

The conventional full-parameter fine-tuning paradigm (i.e., updating all parameters) is computationally expensive for fine-tuning LLMs on edge devices/servers. Moreover, for distributed learning paradigms such as FL and SL, full-parameter fine-tuning also incurs substantial communication overhead associated stemming from model aggregation, hindering the democratization of distributed learning for LLM fine-tuning paradigm. To tackle these challenges, parameter-efficient fine-tuning (PEFT) techniques are a promising solution. There are several representative parameter-efficient fine-tuning methods for LLM, including Adapter Tuning~\cite{houlsby2019parameter, pfeiffer2020adapterfusion, karimi2021compacter}, Prompt Tuning~\cite{li2023prompt}, and Low-rank Adaptation (LoRA)~\cite{hu2021lora, sheng2023s}. Adapter tuning inserts well-designed adapter modules between layers for training, while prompt tuning adds tunable prefix markers. LoRA decomposes incremental matrix of attention weight into low-rank matrices for model updating. The underlying principle of these methods is to train a small fraction of parameters, typically less than 1$\%$ of the original parameters, thus significantly reducing the number of trainable parameters.

Despite our framework can support many PEFT techniques such as Adapter Tuning~\cite{houlsby2019parameter, pfeiffer2020adapterfusion, karimi2021compacter}, Prefix-Tuning~\cite{li2021prefix}, and P-Tuning~\cite{liu2021p}, we prefer to employ LoRA~\cite{hu2021lora} since it requires few trainable parameters for adaptation and does not introduce additional inference latency. LoRA is built on the insights that over-parametrized models essentially reside on a low intrinsic dimension~\cite{aghajanyan2021intrinsic}, and thus, can be characterized with the reduced number of trainable parameters. Specifically, Let $\mathbf{W} \in \mathbb{R}^{d \times m}$ denote one trainable weight matrix of LLM, its corresponding model update is represented as $\mathbf{W} + \Delta \mathbf{W} = \mathbf{W} + \mathbf{A}\mathbf{B}$, where $\mathbf{A} \in \mathbb{R}^{d \times r}$, $\mathbf{B} \in \mathbb{R}^{r \times m}$ are decomposition matrices, and $r \ll \min (d,m)$. In this way, the number of trainable parameters $\bf{W}$ could be less than $1\%$ compared to the full-parameter fine-tuning, improving computing and communication efficiency. Therefore, we deploy LoRA fine-tuning technique in SplitLoRA framework.

\subsection{Split Learning and Large Language Models}

Recently, there have been several preliminary papers exploring FL LLM PEFT paradigms~\cite{fedllm-position,fan2023fate,federatedscopellm,fedit}. Even though these frameworks have adopted various PEFT methods for LLM fine-tuning, such as Adapter~\cite{houlsby2019parameter, pfeiffer2020adapterfusion, karimi2021compacter} and LoRA~\cite{hu2021lora, sheng2023s}, however, LLM fine-tuning is still computational overload for resource-constrained computing entities.
Unfortunately, although SL  holds promise in addressing this issue, research on the SL LLM fine-tuning paradigm is still volid.

\begin{figure}[t]
    \centering
    \includegraphics[width=0.5\textwidth]{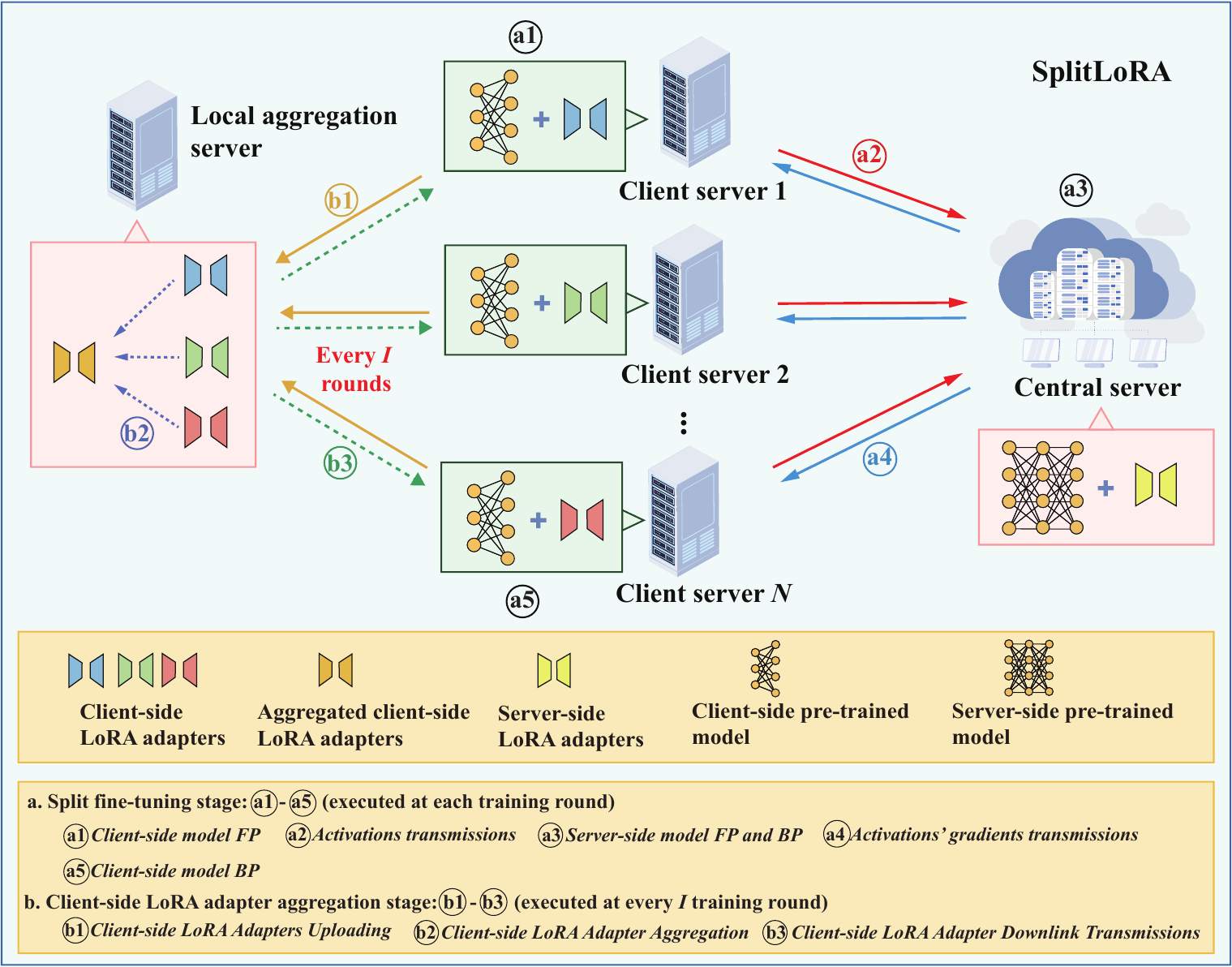}
    \vspace{0.01cm}
    \caption{Overview of proposed SplitLoRA framework.}
    \label{fig:SplitLoRA_fram}
\end{figure}

\section{SplitLoRA Framework}\label{spl_llm}

In this section, we first present the system model and then overview the training procedures of SplitLoRA.

\subsection{System Model}\label{Sys_Model}

In this section, we introduce the system model of SplitLoRA to provide a foundation for the following sections. As illustrated in Fig.~\ref{fig:SplitLoRA_fram}, we consider a typical scenario of SplitLoRA, which consists of three fundamental components:

\begin{itemize}
\item \textbf{Client server:} We assume that each client server has sufficient computing capability to execute the forward propagation (FP) and back-propagation (BP) of the client-side pre-trained model. We denote  $\mathcal{N} = \left\{ {1,2,...,N} \right\}$ as the set of participating client servers, where $N$ is the number of client servers. Each client server $i$ has its local dataset ${\mathcal{D}_i} = \left\{ {{{\bf{x}}_{i,k}},{y_{i,k}}} \right\}_{k = 1}^{{D_i}}$ with ${D_i}$ data samples, where ${{\bf{x}}_{i,k}}$  and ${{{y}}_{i,k}} $ represent the $k$-th input data and its corresponding label. The client-side pre-trained model of each client server is denoted as ${{\bf{W}}_{c}}$. ${\mathcal{R}}_{c,i} = \left\{ {{\bf{A}}_{c,i}^1, {\bf{B}}_{c,i}^1, {\bf{A}}_{c,i}^2, {\bf{B}}_{c,i}^2, ..., {\bf{A}}_{c,i}^{M_c}, {\bf{B}}_{c,i}^{M_c}} \right\}$ represents the set of trainable LoRA adapters of the client-side pre-trained model for client server $i$, where ${\bf{A}}_{c,i}^n$ and ${\bf{B}}_{c,i}^n$ denote the decomposition matrices of $n$-th LoRA adapter of client server $i$ and $M_c$ is total number of trainable LoRA adapters of client-side pre-trained model.

\item \textbf{Central server:} The central server is a powerful computing entity responsible for performing server-side pre-trained model fine-tuning. The server-side pre-trained model is denoted by ${{\bf{W}}_{s}}$. ${\mathcal{R}}_s = \left\{ {{\bf{A}}_{s}^1, {\bf{B}}_{s}^1, {\bf{A}}_{s}^2, {\bf{B}}_{s}^2, ..., {\bf{A}}_s^{M_s}, {\bf{B}}_s^{M_s}} \right\}$ represents the set of trainable LoRA adapters of the server-side pre-trained model, where ${\bf{A}}_{s}^m$ and ${\bf{B}}_{s}^m$ denote the decomposition matrices of $m$-th LoRA adapter and $M_s$ is total number of trainable LoRA adapters of server-side pre-trained model.

% Furthermore, the client server also takes charge of collecting important network information, such as the device computing capabilities and channel states, to support optimization decisions.

\item \textbf{Local aggregation server:} The local aggregation server takes charge of synchronizing client-side LoRA adapters, periodically aggregating them from all participating client servers. Due to privacy concerns, fed and central servers usually do not belong to the identical party since a malicious server could reconstruct the raw data by obtaining client-side pre-trained models and activations~\cite{pasquini2021unleashing}. 
\end{itemize}

% For edge device $i$, the activation obtained from ${{{\bf{x}}_{i,k}}}$ is represented as ${{\bf{a}}_{i,k}} = \varphi\left( {{{\bf{x}}_{i,k}};{{\bf{w}}_{c,i}}} \right)$, where $\varphi\left( {{{x}};{{w}}} \right)$ maps the relationship between input data ${{x}}$ and its predicted value given model parameter ${{w}}$.  Similarly, based on activation ${{{\bf{a}}_{i,k}}}$, we denote its corresponding predicted value as ${{\hat y}_{i,k}} = \varphi\left( {{{\bf{a}}_{i,k}};{{\bf{w}}_{s,i}}} \right)$. 

% The global model is denoted by ${\bf{w}} = \left[ {{{\bf{w}}_{s,i}};{{\bf{w}}_{c,i}}} \right] $. For edge device $i$, local loss function is represented as  ${f_i}\left( {\bf{w}} \right) = {\mathbb{E}_{{\xi _i} \sim {\mathcal{D}_i}}}[{F_i}\left( {{\bf{w}};{\xi _i}} \right)]$

The global pre-trained model is denoted by ${\bf{W}} = \left[ {{{\bf{W}}_{s}};{{\bf{W}}_{c}}} \right] $. The objective of SFL is to find the optimal LoRA adapter configuration ${\mathcal{R}}^*_{c, i}$ and ${\mathcal{R}}^*_s$ that achieves good performance across all participating devices, which can be formulated as minimizing the following learning objective:
\begin{align}\label{minimiaze_loss_function}
\mathop {\min }\limits_{{{\cal R}_{c,i}},{{\cal R}_s}} \sum\limits_{i = 1}^N {\frac{{{\left| {\cal D}_i \right|}}}{\left| \cal D \right|}{f_i}} ({\bf{W}}|{{\cal R}_{c,i}},{{\cal R}_s})
\end{align}
where $ \mathcal{D} = {\textstyle \bigcup_{i=1}^{N} \mathcal{D}_{i}}$ is total dataset size and ${f_i}\left( {\bf{W}} \vert {\mathcal{R}}_{c, i}, {\mathcal{R}}_s \right)$ denotes the local loss function of client server $i$ over local dataset $\mathcal{D}_i$.

\subsection{The SplitLoRA Framework}\label{subsec_adapt}
This section provides a detailed overview of the workflow of the proposed SplitLoRA framework. The distinctive feature of SplitLoRA lies in the integration of the advantages of model splitting from SL and parallel training from FL. Moreover, SplitLoRA employs the state-of-the-art fine-tuning technique, LoRA, to further enhance the model training efficiency.

Before model training begins, the central server initializes the ML model and partitions it into client-side and server-side pre-trained models. Afterward, SplitLoRA conducts model fine-tuning over $I$ consecutive training rounds, after that the client-side LoRA adapters aggregation is performed. This process loops until the model converges. The training process of SplitLoRA comprises two primary stages: split fine-tuning and client-side LoRA adapters aggregation. The split fine-tuning is executed in each training round, while client-side LoRA adapters aggregation occurs every $I$ round. As illustrated in Fig.~\ref{fig:SplitLoRA_fram}, for a training round $t \in \mathcal{T} = \left\{ {1,2,...,T} \right\}$, the training process of SplitLoRA is detailed as follows.

\textit{a. Split Fine-Tuning Stage:} The split fine-tuning stage involves client-side and server-side fine-tuning for participating client servers and a central server in each training round, which consists of the following five steps.

\textit{a1) Client-side Model Forward Propagation:} In this step, all participating client servers conduct the FP of client-side pre-trained model in parallel. Specifically, each client server $i$ randomly selects a mini-batch ${\mathcal{B}_i} \subseteq {\mathcal{D}_i}$ with $b$ data samples from its local dataset for fine-tuning client-side pre-trained model. The input data and corresponding label of mini-batch in the training round $t$ are denoted by ${{\bf{x}}^t_i}$ and ${{\bf{y}}^t_i}$, respectively. ${\mathcal{R}}^t_{c,i} = \left\{ {{\bf{A}}_{c,i}^{1,t}, {\bf{B}}_{c,i}^{1,t}, {\bf{A}}_{c,i}^{2,t}, {\bf{B}}_{c,i}^{2,t}, ..., {\bf{A}}_{c,i}^{M_c,t}, {\bf{B}}_{c,i}^{M_c,t}} \right\}$ represents the set of trainable LoRA adapters of the client-side pre-trained model for client server $i$ at the $t$-th training round. After feeding a mini-batch of data into the client-side pre-trained model, activations are generated at the cut layer. The activations of client server $i$ are given by
\begin{align}\label{stage_1_1}
{\bf{s}}_i^t = \varphi \left( {{\bf W}_{c} \vert {{\cal R}^{t-1}_{c,i}},{\bf{x}}_i^t} \right),
\end{align}
where $\varphi\left( {{{w}} \vert {r} ,{{x}}} \right)$ denotes the mapping relationship between input data ${{x}}$ and its predicted value given model parameter ${{w}}$ and trainable LoRA adapters set $r$.

\textit{a2) Activations Transmissions:} 
After completing the FP of client-side pre-trained models, all participating client servers transmit their respective activations and corresponding labels to the central server (usually over wireless channels). The collected activations from participating client servers are then utilized to fuel server-side model fine-tuning.

\textit{a3) Server-side Model Forward Propagation and Back-propagation:} After receiving activations from participating client servers, the central server feeds these activations into the server-side pre-trained model to execute the server-side FP. ${\mathcal{R}}^t_{s} = \left\{ {{\bf{A}}_{s}^{1,t}, {\bf{B}}_{s}^{1,t}, {\bf{A}}_{s}^{2,t}, {\bf{B}}_{s}^{2,t}, ..., {\bf{A}}_{s}^{M_s,t}, {\bf{B}}_{s}^{M_s,t}} \right\}$ represents the set of trainable LoRA adapters of the server-side pre-trained model at the $t$-th training round. The concatenated activation matrix ${{\bf{S}}^t}$ is denoted by ${{\bf{S}}^t} = \left[ {{{\bf{s}}_1^t};{{\bf{s}}_2^t};...;{{\bf{s}}_N^t}} \right]$. Thus, the predicted value is expressed as
\begin{align}\label{stage_1_3}
{\bf{\hat y}}^t_i = \varphi \left( {{\bf W}_{s} \vert {{\cal R}^{t-1}_{s}},{\bf{S}}^t} \right), 
\end{align}
The predicted value and labels are utilized to calculate loss function value and further derive the server-side LoRA adapters' gradients. Therefore, the $m$-th server-side LoRA adapters can be updated through
\begin{align}\label{stage_5_2}
{\bf{A}}_s^{m,t} \leftarrow {\bf{A}}_s^{m,t - 1} - \gamma_s {\bf{G}}_{{A}, s}^{m,t},\\
{\bf{B}}_s^{m,t} \leftarrow {\bf{B}}_s^{m,t - 1} - \gamma_s {\bf{G}}_{{B}, s}^{m,t},
\end{align}
where ${\bf{G}}_{{A}, s}^{m,t}$ and  ${\bf{G}}_{{B}, s}^{m,t}$ denote the gradients of the decomposition matrices $\bf A$ and $\bf B$ of $m$-th LoRA adapter of server-side pre-trained model, and $\gamma_s$ is the server-side learning rate.

\textit{a4) Activations' Gradients Transmissions:} After the server-side BP is completed, the central server transmits the activations' gradients to its corresponding participating client servers.

\textit{a5) Client-side Model Back-propagation:} In this step, each client server fine-tunes its client-side pre-trained model based on the received activations' gradients. For client server $i$, the $n$-th client-side LoRA adapters is updated through
\begin{align}\label{client_side_update}
{\bf{A}}_{c,i}^{n,t} \leftarrow {\bf{A}}_{c,i}^{n,t - 1} - \gamma_c {\bf{G}}_{{A}, {c,i}}^{n,t},\\
{\bf{B}}_{c,i}^{n,t} \leftarrow {\bf{B}}_{c,i}^{n,t - 1} - \gamma_c {\bf{G}}_{{B}, {c,i}}^{n,t},
\end{align}
where ${\bf{G}}_{{A}, c, i}^{n,t}$ and  ${\bf{G}}_{{B}, c, i}^{n,t}$ denote the gradients of the decomposition matrices $\bf A$ and $\bf B$ of $n$-th LoRA adapter of server-side pre-trained model, and $\gamma_c$ is the client-side learning rate.

\textit{b. Client-side LoRA Adapter Aggregation Stage:} The client-side LoRA adapter aggregation stage primarily focuses on aggregating client-side LoRA adapters on the local aggregation server, which is executed every $I$ training round. This stage consists of the following three steps.

\textit{b1) Client-side LoRA Adapters Uploading:} In this step, each participating client server sends its client-side LoRA adapters to the local aggregation server over the wireless/wired links.

\textit{b2) Client-side LoRA Adapter Aggregation:} The local aggregation server aggregated the received client-side LoRA adapters into aggregated LoRA adapters. The decomposition matrices $\bf A$ and $\bf B$ of $n$-th client-side LoRA adapter are aggregated separately as follows
\begin{align}\label{h_c_define}
{\bf{A}}_{{c}}^{n,t} = \sum\limits_{i = 1}^N {\frac{{{\left| {\cal D}_i \right|}}}{\left| \cal D \right|}} {\bf{A}}_{ {c,i}}^{n,t},\\
{\bf{B}}_{{c}}^{n,t} = \sum\limits_{i = 1}^N {\frac{{{\left| {\cal D}_i \right|}}}{\left| \cal D \right|}} {\bf{B}}_{ {c,i}}^{n,t}.
\end{align}

\textit{b3) Client-side LoRA Adapter Downlink Transmissions:} After completing client-side LoRA adapter aggregation, the local aggregation server sends aggregated client-side LoRA adapter to the participating client servers. Afterward, each client server utilizes the received aggregated LoRA adapter as the initial LoRA configuration for the next training round.

The SplitLoRA training framework is outlined in~\textbf{Algorithm~\ref{AdaptSFL_procedure}}.

\begin{algorithm}[t]
	%\textsl{}\setstretch{1.8}
	\renewcommand{\algorithmicrequire}{\textbf{Input:}}
	\renewcommand{\algorithmicensure}{\textbf{Output:}}
	\caption{The SplitLoRA Training Framework}\label{AdaptSFL_procedure}
	\begin{algorithmic}[1]
 \REQUIRE   $b$, $I$, $\gamma_c$, $\gamma_s$, $E$, ${\cal N}$, ${\mathcal{R}}^0_{s}$, ${\mathcal{R}}^0_{c}$, ${\bf W}_c$, ${\bf W}_s$ and $\cal{D}$.
		\ENSURE ${\mathcal{R}}^*_{s}$, ${\mathcal{R}}^*_{c}$. 
		\STATE Initialization: ${{{\mathcal{R}}}_{c,i}^{{0}}} \leftarrow {{{\mathcal{R}}}_c^{{0}}}$.
          \FOR {$t=1, 2, ..., T$}
          \STATE
          \STATE /** {Runs} {on} {edge} {servers} **/
          \FORALL {client server ${i \in \,{\cal N}}$ in parallel}
            \STATE  ${\bf{s}}_i^t = \varphi \left( {{\bf W}_{c} \vert {{\cal R}^{t-1}_{c,i}},{\bf{x}}_i^t} \right)$
            \STATE Send $\left( {{{\bf{s}}^t_i},{\bf{y}}_i^t} \right)$ to the central server
          \ENDFOR

	   \STATE
          \STATE /** {Runs} {on} {central} {server} **/
          \STATE ${{\bf{S}}^t} = \left[ {{{\bf{s}}_1^t};{{\bf{s}}_2^t};...;{{\bf{s}}_N^t}} \right]$
          \STATE${\bf{\hat y}}^t_i = \varphi \left( {{\bf W}_{s} \vert {{\cal R}^{t-1}_{s}},{\bf{S}}^t} \right)$
          \STATE Calculate loss function value ${f_i}\left( {\bf{W}}^{t-1} \vert {\mathcal{R}}_{c, i}, {\mathcal{R}}_s \right)$
          \STATE ${\bf{A}}_s^{m,t} \leftarrow {\bf{A}}_s^{m,t - 1} - \gamma_s {\bf{G}}_{{A}, s}^{m,t}$ 
          \STATE ${\bf{B}}_s^{m,t} \leftarrow {\bf{B}}_s^{m,t - 1} - \gamma_s {\bf{G}}_{{B}, s}^{m,t}$
          \STATE Send activations' gradients  to corresponding client servers

	   \STATE
          \STATE /** {Runs} {on} {edge} {servers} **/
         \FORALL {edge device ${i \in \,{\cal N}}$ in parallel}
           \STATE ${\bf{A}}_{c,i}^{n,t} \leftarrow {\bf{A}}_{c,i}^{n,t - 1} - \gamma_c {\bf{G}}_{{A}, {c,i}}^{n,t}$
            \STATE ${\bf{B}}_{c,i}^{n,t} \leftarrow {\bf{B}}_{c,i}^{n,t - 1} - \gamma_c {\bf{G}}_{{B}, {c,i}}^{n,t}$
        \ENDFOR
        \STATE
        \STATE /** {Runs} {on} the {fed} {server} **/
        \IF{$T$ mod $I$ $=0$}
          \STATE ${\bf{A}}_{{c}}^{n,t} = \sum\limits_{i = 1}^N {\frac{{{\left| {\cal D}_i \right|}}}{\left| \cal D \right|}} {\bf{A}}_{ {c,i}}^{n,t}$
          \STATE ${\bf{B}}_{{c}}^{n,t} = \sum\limits_{i = 1}^N {\frac{{{\left| {\cal D}_i \right|}}}{\left| \cal D \right|}} {\bf{B}}_{ {c,i}}^{n,t}$
        \ENDIF
        \ENDFOR 
	\end{algorithmic}  
\end{algorithm}

\begin{figure}[t]
\centering  
\subfigure[GPT2-S.]{
\label{Fig.sub.1}
\includegraphics[width=4.2cm,height = 3.8cm]{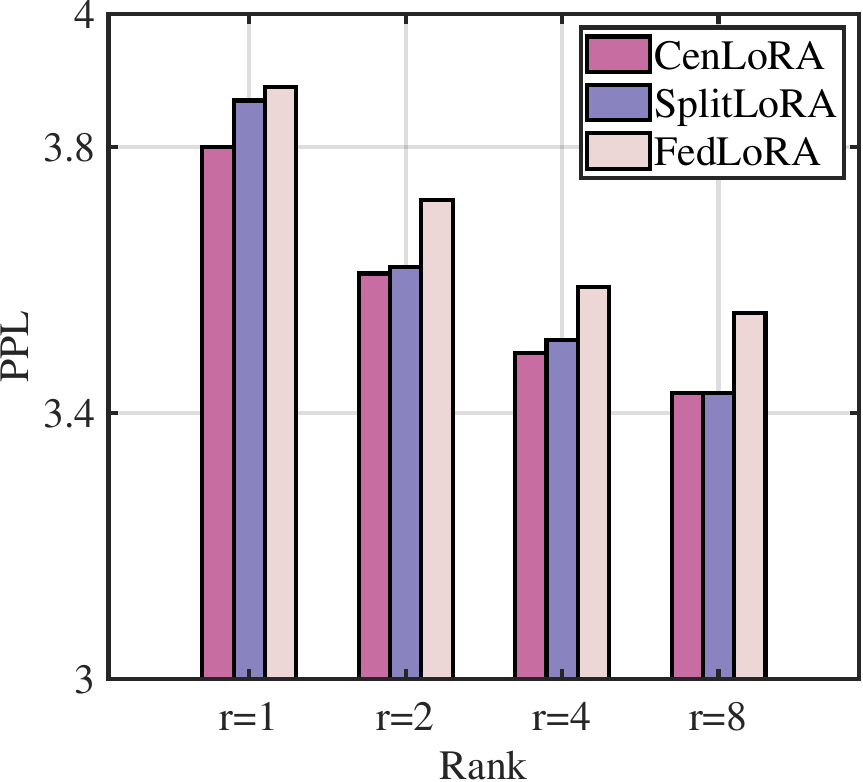}}
\subfigure[GPT2-M.]{
\label{Fig.sub.2}
\includegraphics[width=4.2cm,height = 3.8cm]{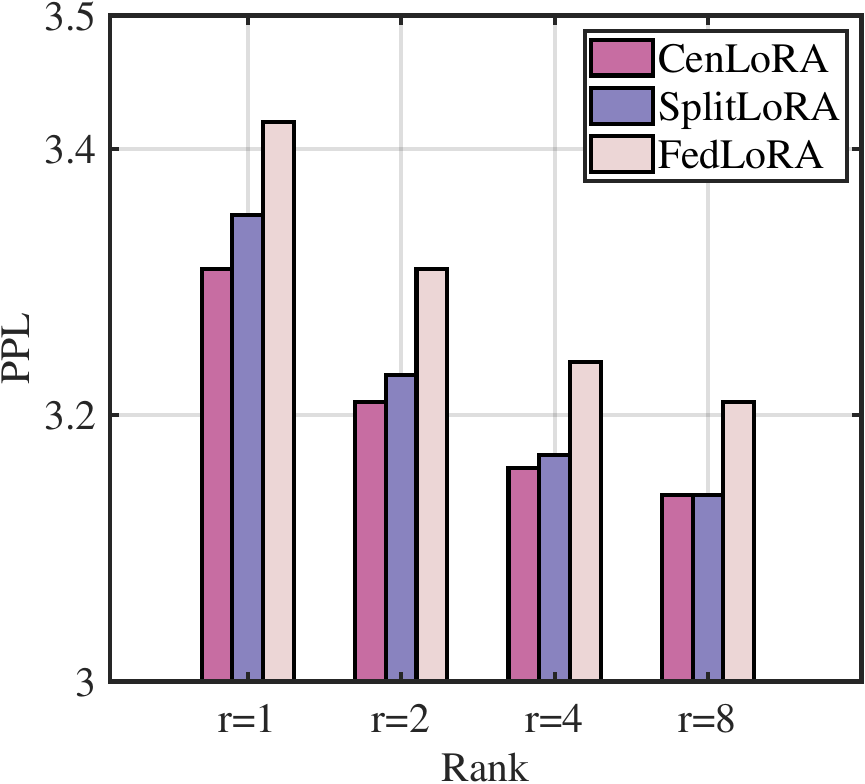}}
\caption{The converged accuracy for GPT2-S and GPT2-M models, where Perplexity (PPL) is a metric used to measure how well LLMs predict a sample, with lower PPL indicating better predictive performance.}
\label{PPL_compare}
\end{figure}

% \begin{figure}[t]
% \centering  
% \figure[GPT2-S.]{
% \label{Fig.sub.1}
% \includegraphics[width=4.2cm,height = 3.8cm]{PPL_GPT2-S.pdf}}
% \subfigure[GPT2-M.]
% \end{figure}

% \begin{figure}[t]
%     \centering
%     \includegraphics[width=0.21\textwidth]{experimental_set.jpg}
%     \caption{Experimental implementation.}
%     \label{experimental_set}
% \end{figure}

\section{Performance Evaluation}\label{experiment}

This section provides experimental results to evaluate the training performance of the SplitLoRA framework in terms of converged accuracy, convergence rate, and resource efficiency.

% This section provides extensive evaluations to demonstrate SplitLoRA's advantages in overall performance, convergence rate, and resource efficiency.
% , and trainable parameters.
% We first describe basic setups. Then,we evaluate the overall performance of SplitLoRA and conducted experiments to illustrate the advantages of the SplitLoRA design.

\subsection{Experimental Setup}

% \subsubsection{Implementation}
{\bf{Hardware:}} We implement SplitLoRA using Python 3.7 and PyTorch 1.7.1., and train it with the NVIDIA GeForce RTX 3090 GPUs.

% \needrev{SplitLoRA was configured with three clients, each holding the first three layers of the GPT-2 model, while the server held the remaining 9 layers (GPT2-S) or 21 layers (GPT2-M).}
% We configured CenLoRA to hold the entire E2E dataset for training.
% FedLoRA was set up with three clients, each possessing a complete GPT-2 model and one-third of the E2E dataset as local data.
% SplitLoRA was configured with three clients, each holding the first three layers of the GPT-2 model and one-third of the E2E dataset as local data, while the server held the remaining 9 layers (GPT-2-S) or 21 layers (GPT-2-M).
% We evaluate the performance of SplitLoRA for NLG tasks on E2E dataset.
% We train and test the model on the E2E dataset using Natural Language Generation (NLG) tasks.
% NLG tasks are aiming to transform various forms of inputs into human-readable language.
% The E2E dataset is a benchmark dataset used for evaluating and researching NLG tasks that generate natural language text.

{\bf{Dataset and task:}} We evaluate the training performance of SplitLoRA for natural language generation (NLG) tasks on E2E dataset~\cite{novikova2017e2e}. The E2E dataset consists of around 42000 training, 4600 validation, and 4600 test examples from the restaurant domain.

{\bf{Model:}} We employ well-known GPT2 small(GPT-S) and GPT2 medium (GPT2-M) models.
% GPT2 is a large language model developed by OpenAI, based on the Transformer architecture, capable of generating high-quality natural language text.
GPT2-S is the smallest version in the GPT2 series
with 124 million parameters and 12-layer Transformer encoders, while GPT2-M is the medium-sized version of GPT2, incorporating 355 million parameters, and 24-layer Transformer encoders.

{\bf{Experimental setup:}} In the simulations, we deploy $N=3$ client servers by default unless specified otherwise. The client-side model of each client server has the first three Transformer layers of the GPT-2 model, while server-side model of central server holds the remaining 9 Transformer layers (GPT2-S) or 21 layers (GPT2-M). The computing capability of each client server is 35.6 TFLOTS (peak performance of one NVIDIA RTX 3090), while the central server's computing capability is set to 284.8 TFLOTS. {The communication rate between the client server and the local aggregation server is set to 300 Mbps, and the communication rate between the client server and the central server is set to 600Mbps. } We set the mini-batch size, learning rate, and maximum sequence length to 8, 0.0002, and 512 for the GPT2-S model, and 4, 0.0002, and 512 for the GPT2-M model. The rank of the LoRA adapter is set to ${r} = \left\{ {1,2,4,8} \right\}$.

{\bf{Benchmarks:}} To investigate the
advantages of SplitLoRA, we compare it with two canonical benchmarks: (1) \textbf{Centralized LoRA (CenLoRA):} Client server collects raw data from other participating servers for full model fine-tuning with LoRA adapters. (2) \textbf{Federated LoRA (FedLoRA):} Each participating client server locally fine-tunes the full model and then transmits the updated LoRA adapters to the central server for adapter aggregation.

%\textbf{Federated LoRA (FedLoRA):} This method involves training on multiple edge client devices with sufficient computational power and local datasets, and then aggregating the client models to obtain a global model.

%\textbf{Federated LoRA (FedLoRA):} This method combines the principles of federated learning with LoRA, enabling on-device model personalization and learning across decentralized data sources while maintaining data localization. 

%By training low-rank modules on individual devices and aggregating the updates, \needrev{it substantially reduces communication costs and bolsters privacy.}

\begin{table}[t]
\renewcommand{\arraystretch}{1.2}
\begin{tabular}{|p{2cm}|p{0.7cm}|p{0.7cm}|p{1cm}|p{1.1cm}|p{0.7cm}|}
\hline
\multicolumn{1}{|c|}{GPT2-S}    & BLEU   & NIST   & METEOR & ROUGE\_L & CIDEr  \\ \hline
CenLoRA(r=1) & 67.95 & 8.6973 &\makecell[c]{ 0.4421} & \makecell[c]{68.96}   & 2.3412 \\ 
CenLoRA(r=2) & 68.49 & 8.7481 & \makecell[c]{0.4491} & \makecell[c]{68.70}   & 2.3952 \\ 
CenLoRA(r=4) & 69.41 & 8.7824 & \makecell[c]{0.4610} & \makecell[c]{70.70}   & 2.4713 \\ 
CenLoRA(r=8) & 69.37 & 8.7735 & \makecell[c]{0.4624} & \makecell[c]{70.96}   & 2.4572 \\ \hline
SplitLoRA(r=1)     & 67.18 & 8.6601 & \makecell[c]{0.4416} & \makecell[c]{67.71}   & 2.3255 \\ 
SplitLoRA(r=2)     & 66.86 & 8.5667 & \makecell[c]{0.4515} & \makecell[c]{68.50}   & 2.3358 \\ 
SplitLoRA(r=4)     & 68.79 & 8.7259 & \makecell[c]{0.4572} & \makecell[c]{69.84}   & 2.4411 \\ 
SplitLoRA(r=8)     & 68.76 & 8.6931 & \makecell[c]{0.4588} & \makecell[c]{70.17}   & 2.4165 \\ \hline
FedLoRA(r=1)       & 65.66 & 8.4123 & \makecell[c]{0.4265} & \makecell[c]{67.68}   & 2.1921 \\ 
FedLoRA(r=2)       & 67.24 & 8.6055 & \makecell[c]{0.4398} & \makecell[c]{69.33}   & 2.3025 \\ 
FedLoRA(r=4)       & 67.73 & 8.6148 & \makecell[c]{0.4494} & \makecell[c]{68.59}   & 2.3817 \\ 
FedLoRA(r=8)       & 68.39 & 8.6745 & \makecell[c]{0.4590} & \makecell[c]{70.24}   & 2.4450 \\ \hline
\end{tabular}
\vspace{3mm}
\caption{The comparison of converged accuracy on GPT2-S for E2E NLG challenge.}
\label{tab:gtp2-s}
\end{table}

\begin{table}[t]
\renewcommand{\arraystretch}{1.2}
\begin{tabular}{|p{2cm}|p{0.7cm}|p{0.7cm}|p{1cm}|p{1.1cm}|p{0.7cm}|}
\hline
\multicolumn{1}{|c|}{GPT2-M}& BLEU   & NIST   & METEOR & ROUGE\_L & CIDEr  \\ \hline
CenLoRA(r=1)             & 69.86 & 8.7679 & \makecell[c]{0.4650} & \makecell[c]{71.20}   & 2.5028 \\ 
CenLoRA(r=2)             & 69.97 & 8.7787 & \makecell[c]{0.4663} & \makecell[c]{71.56}   & 2.5029 \\ 
CenLoRA(r=4)             & 69.78 & 8.7820 & \makecell[c]{0.4667} & \makecell[c]{71.62}   & 2.5301 \\ 
CenLoRA(r=8)             & 70.57 & 8.8557 & \makecell[c]{0.4688} & \makecell[c]{72.17}   & 2.5405 \\ \hline
SplitLoRA(r=1)              & 70.26 & 8.8274 & \makecell[c]{0.4664} & \makecell[c]{71.73}   & 2.5267 \\ 
SplitLoRA(r=2)              & 70.04 & 8.8031 & \makecell[c]{0.4670} & \makecell[c]{71.68 }  & 2.5233 \\ 
SplitLoRA(r=4)                & 70.09 & 8.8075 & \makecell[c]{0.4667} & \makecell[c]{71.60}   & 2.5370 \\ 
SplitLoRA(r=8)                & 69.18 & 8.7189 & \makecell[c]{0.4631} & \makecell[c]{71.30}   & 2.5156 \\ \hline
FedLoRA(r=1)         & 67.02 & 8.6467 & \makecell[c]{0.4484} & \makecell[c]{68.06}   & 2.3431 \\ 
FedLoRA(r=2)        & 69.64 & 8.7727 & \makecell[c]{0.4633} & \makecell[c]{71.35}   & 2.4900 \\ 
FedLoRA(r=4)         & 69.78 & 8.7836 & \makecell[c]{0.4642} & \makecell[c]{71.87}   & 2.4819 \\ 
FedLoRA(r=8)         & 69.55 & 8.7358 & \makecell[c]{0.4661} & \makecell[c]{71.46 }  & 2.4980\\ \hline
\end{tabular}
\vspace{3mm}
\caption{The comparison of converged accuracy on GPT2-M for E2E NLG challenge.}
\label{tab:gtp2-m}
\end{table}

\subsection{Performance Evaluation}
{\bf{Converged accuracy:}} Fig.~\ref{PPL_compare} compares the training performance of SplitLoRA with other benchmarks on GPT2-S and GPT2-M models for the E2E NLG challenge, with PPL as the performance metric. FedLoRA exhibits lower converged accuracy than SplitLoRA and CenLoRA, with PPL approximately 0.08/0.11 (GPT2-S/GPT2-M) and 0.73/0.09 (GPT2-S/GPT2-M) higher than SplitLoRA and CenLoRA, respectively. The poorer training performance of FedLoRA stems from its fully distributed training paradigm, where the entire model is updated via model aggregation, making it susceptible to model bias caused by data heterogeneity across client servers. SplitLoRA achieves converged accuracy comparable to that of cenLoRA, especially on GPT2-M, where the accuracy difference is less than 0.04. The reason for this is two-fold: One is that server sub-models in SplitLoRA are trained in a centralized manner, making it more resilient to data heterogeneity. The other is that SplitLoRA offloads the major workload to a central server, with only a small portion of LLM affected by client data heterogeneity. It can be seen that PPL of SplitLoRA and CenLoRA becomes closer with the rank $r$ increases. This is because that higher $r$ implies a larger number of trainable parameters, thereby enhancing the fitting capability of the server-side sub-model. This may partially counteract the effect of data heterogeneity, bringing the performance of SplitLoRA closer to that of CenLoRA. The comparison of training performance on other metrics is shown in Table~\ref{tab:gtp2-s} and~\ref{tab:gtp2-m}.

\begin{figure}[t]
\centering  
\subfigure[GPT2-S.]{
\label{fig:SplitLoRA GPT2-S training process}
\includegraphics[width=4.2cm,height = 3.8cm]{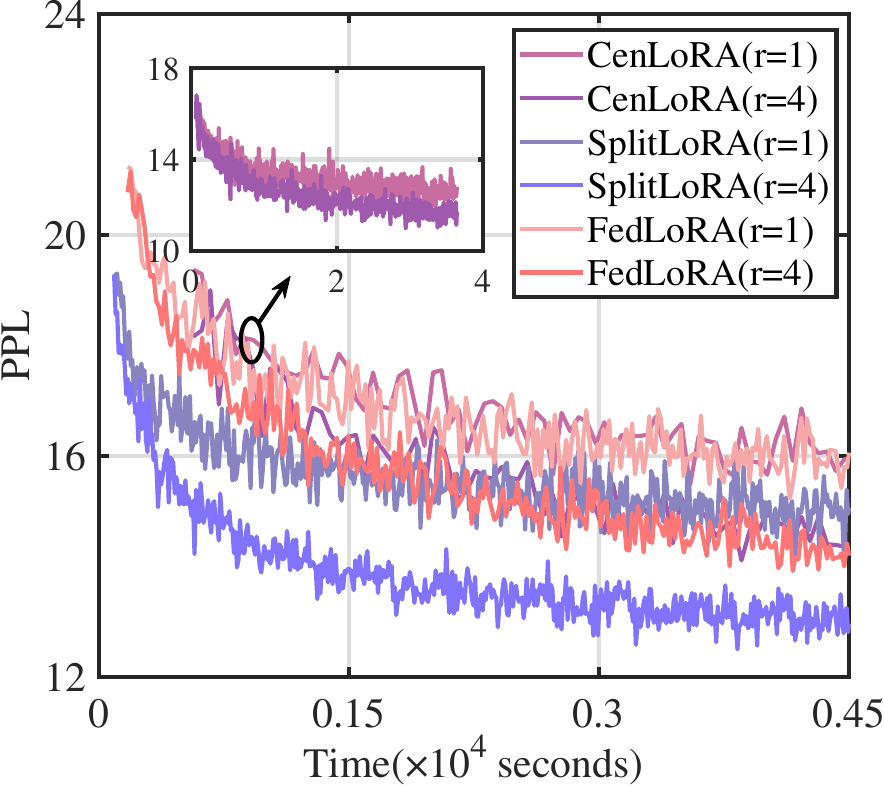}}
\subfigure[GPT2-M.]{
\label{fig:SplitLoRA GPT2-M training process}
\includegraphics[width=4.2cm,height = 3.8cm]{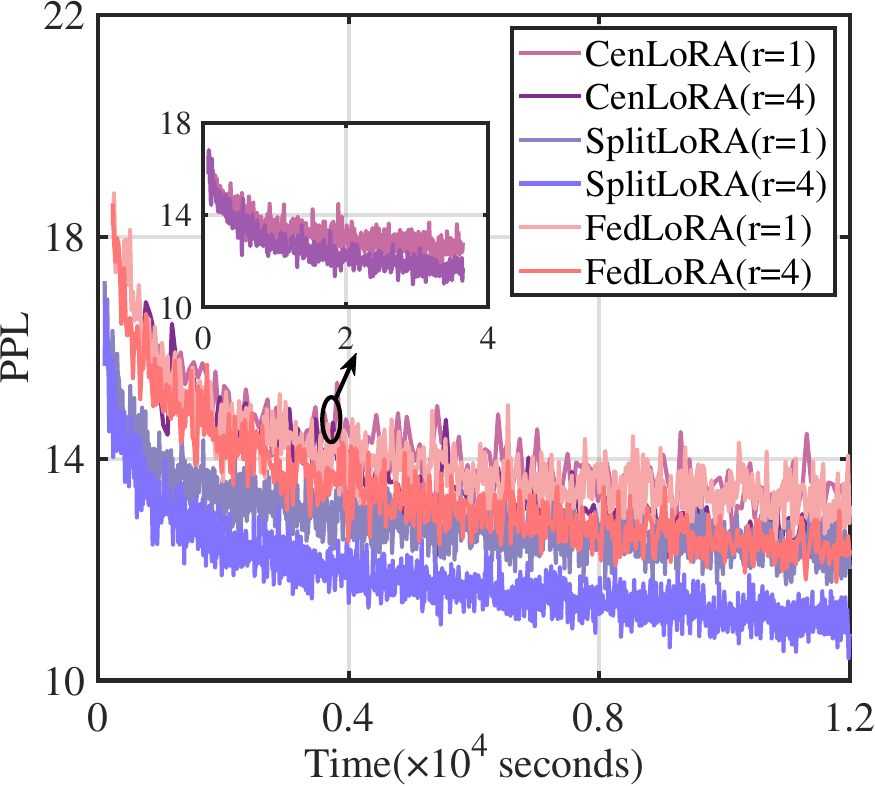}}
\caption{The training performance on GPT2-S and GPT2-M for E2E NLG challenge.}
\label{Fig.The converged accuracy for GPT2-S and GPT2-M models}
\end{figure}

\begin{table}[t]
\renewcommand{\arraystretch}{1.2}
\centering
\begin{tabular}{|p{2cm}|p{1.5cm}|p{1.5cm}|}
\hline
\multicolumn{1}{|c|}{}& \makecell[c]{GPT2-S}   & \makecell[c]{GPT2-M}    \\ \hline
         \makecell[l]{CenLoRA(r=1)} & \makecell[c]{0.031M} & \makecell[c]{0.089M} \\ 
         \makecell[l]{CenLoRA(r=2)} & \makecell[c]{0.062M} & \makecell[c]{0.178M} \\ 
         \makecell[l]{CenLoRA(r=4)} & \makecell[c]{0.124M} & \makecell[c]{0.355M} \\ 
         \makecell[l]{CenLoRA(r=8)} & \makecell[c]{0.248M} & \makecell[c]{0.710M} \\ \hline
         \makecell[l]{SplitLoRA(r=1) } & \makecell[c]{0.008M}  & \makecell[c]{0.011M} \\ 
         \makecell[l]{SplitLoRA(r=2) } & \makecell[c]{0.015M} & \makecell[c]{0.022M} \\ 
         \makecell[l]{SplitLoRA(r=4) } & \makecell[c]{0.031M} & \makecell[c]{0.044M} \\ 
         \makecell[l]{SplitLoRA(r=8) } & \makecell[c]{0.062M} & \makecell[c]{0.088M} \\ \hline
         \makecell[l]{FedLoRA(r=1)} & \makecell[c]{0.031M} & \makecell[c]{0.089M} \\  
         \makecell[l]{FedLoRA(r=2)} & \makecell[c]{0.062M} & \makecell[c]{0.178M} \\  
         \makecell[l]{FedLoRA(r=4)} & \makecell[c]{0.124M} & \makecell[c]{0.355M} \\  
         \makecell[l]{FedLoRA(r=8)} & \makecell[c]{0.248M} & \makecell[c]{0.710M} \\ \hline
\end{tabular}
\vspace{3mm}
\caption{The comparison of the number of trainable parameters on GPT2-S and GPT2-M.}
\label{tab:Trainable_Parameters}
\end{table}

{\bf{Convergence rate:}} Fig.~\ref{fig:SplitLoRA GPT2-S training process} and Fig.~\ref{fig:SplitLoRA GPT2-M training process} show the training performance of SplitLoRA and other benchmarks on GPT2-S and GPT-M for E2E NLG challenge. It is clear that SplitLoRA has an overwhelming superiority in convergence speed over FedLoRA and CenLoRA. The training latency of FedLoRA and CenLoRA for achieving model convergence is approximately 1.7 times and 4.7 times and 2.1 times and 4.8 times that of SplitLoRA on GPT-S and GPT-M, respectively. Compared to CenLoRA, FedLoRA expedites model convergence due to its parallel training paradigm of multiple client servers. SplitLoRA is built on FedLoRA by offloading the major computing workload to more powerful central server, further reducing the time required for achieving model convergence while supporting parallel training across multiple client servers.

% Compared to CenLoRA, FedLoRA improves the convergence speed by parallel computation of multiple clients.
% Compared with FedLoRA, SplitLoRA further enhances overall convergence rate by offloading most of the model training tasks to a high-computation-capacity server, which not only accelerates training but also mitigates the hindrance of client data heterogeneity in distributed training.
%Fig.~\ref{fig:SplitLoRA GPT2-S training process} and Fig.~\ref{fig:SplitLoRA GPT2-M training process} highlight the superiority of SplitLoRA in terms of convergence rate. We set up 3 clients, 1 main server, and 1 federated server. The computing speed of each client is 35.6 TFLOTS, while the server's computing speed is 284.8 TFLOTS. The communication speed is \needrev{300 MB/s. With} the same training configuration, SplitLoRA effectively utilizes the server's high computational power and the client's parallel capabilities to increase the convergence rate. Given the same number of training epochs, SplitLoRA theoretically necessitates only 50\% of the training time compared to FedLoRA, and 20\% of the time compared to CenLoRA.

{\bf{Trainable parameter:}} Table~\ref{tab:Trainable_Parameters} presents the number of trainable parameters on GPT2-S and GPT2-M. SplitLoRA reduces the computing workload on client servers by partitioning the model and retaining fewer Transformer layers locally. In our experimental setup, for the GPT2-S and GPT2-M models, the parameters involved in fine-tuning on client servers in SplitLoRA constitute only one-fourth and one-eighth of the entire model, respectively. This substantial reduction in local computing requirements renders SplitLoRA highly efficient and suitable for deployment in resource-limited environments. In contrast, CenLoRA and FedLoRA require client servers to handle the entire GPT-2 model for fine-tuning, which imposes significant demands on computing capabilities and memory resources. In particular, in 5G and beyond edge computing systems, small base stations and access points with limited capabilities serve as client servers. Therefore, SplitLoRA not only improves computing efficiency but also enhances the flexibility and adaptability of the system to better cope with varying computational and resource-limited scenarios.

\section{Discussions and Future Directions}\label{future_dir}

Previously, we have demonstrated the proposed SplitLoRA framework and its outstanding performance. However, this is not the end of the road, as our framework holds potential for further refinement to adapt diverse scenarios and applications. This section delves into the emerging challenges and intriguing directions for future exploration.

\subsection{Model Splitting in SplitLoRA}\label{sec:model_split}
For the SL LLM fine-tuning framework, research on the model splitting is of paramount importance. The cut layer plays a pivotal role in SL, determining the division of computing workload and the volume of data exchanged between client servers/devices and the central server~\cite{lin2023split,lin2023pushing}. LLMs typically consist of billions of parameters, with significant variations in the output size between different layers~\cite{lin2023pushing,qu2024trimcaching,lin2023split}. Therefore, selecting the cut layer can control the data volume transmitted to the central server. Additionally, the cut layer also has impacts on the division of computing workload between client servers/devices and the central server, e.g., deeper cut layers entail a larger computing workload offloaded to the central server. The above observations reveal that the cut layer selection has a dual impact on both communication and computing efficiency, underscoring its significance in SL LLM fine-tuning.

\subsection{Heterogeneous Configuration in SplitLoRA}
\label{sec:future_hetero_prefer}

In practice, the available resources among different client servers/devices vary greatly, and the resources allocated for training may change during runtime, depending on how the on-demand running procedure prioritizes the resource allocation. More importantly, due to the large-scale parameters of LLM, the resource heterogeneity of client servers/devices can lead to vastly different training times, causing a severe straggler problem in model aggregation. Therefore, it is essential to configure heterogeneous fine-tuning module structures to client servers/devices with heterogeneous resources. For instance, in SplitLoRA, joint selecting cut layers and ranks of LoRA adapters is of paramount importance. Recalling section~\ref{sec:model_split}, cut layer selection determines the division of computing workload and the volume of data exchanged between client servers/devices and the central server. Therefore, it is an intuitive idea to select distinct cut layers for client servers/devices with heterogeneous resources. Moreover, the rank of client-side LoRA adapters simultaneously affects the computing burden for client-side LLM fine-tuning and the communication overhead required for client-side LoRA adapter aggregation. Thus, for SplitLoRA, how to jointly select cut layers and ranks of LoRA adapters according to heterogeneous resources is a worthwhile endeavor.

\subsection{Efficiency in SplitLoRA}

SplitLoRA typically involves multiple organizations or data centers, each with sufficient computing resources. In this case, seamless training of SplitLoRA is feasible, as each entity conducting local LLM fine-tuning may possess hardware capabilities similar to or exceeding those of the NVIDIA 3090 GPU. This setting allows for more straightforward collaborative LLM fine-tuning, facilitating the handling of larger model parameters and more complex training routines due to the higher computing resources available. Conversely, the practical implementation of SplitLoRA usually necessitates leveraging private data residing on edge devices (e.g., smartphones and laptop) with lower computing and storage resources than data centers for LLM fine-tuning. Training an LLM with billions of parameters across edge devices poses significant barriers due to the limited storage and computing capabilities. Therefore, efficiency becomes even more critical since the LLMs are usually much larger than conventional models used in previous SL literature~\cite{lin2023split,vepakomma2018split,lin2023efficient}.

Fortunately, recent advancements in model compression, such as knowledge distillation~\cite{wang2023can} and pruning~\cite{xia2023sheared} and quantization~\cite{dettmers2024qlora} techniques, offer promising solutions to reduce model size and computational complexity without significantly compromising performance. These techniques have the potential to enable the deployment of smaller, more efficient versions of LLMs. Therefore, designing efficient model compression and quantization techniques to achieve a more storage, communication, and computation-efficient SL LLM fine-tuning framework is warranted.

\subsection{Privacy Preservation in SplitLoRA}

Deep learning models, those of substantial size, are capable of memorizing training data, which could raise privacy concerns~\cite{nasr2019comprehensive,gupta2022recovering}. This risk is particularly severe in LLM fine-tuning due to their powerful capability, which might inadvertently memorize and potentially expose more detailed information~\cite{carlini2021extracting}. Therefore, it is crucial to design efficient privacy-preserving mechanisms/strategies that guarantee the training performance and effectiveness of SplitLoRA without compromising individual privacy. Two promising directions are worth exploring in the future: one potential avenue is to extend the SplitLoRA framework to a U-shaped paradigm. The SplitLoRA ultimately gets the final inference results in the cloud, however, in the context where training tasks are highly privacy-sensitive (e.g., medical image analysis), this can lead to data privacy leakage of edge devices/servers~\cite{ni2023eavesdropping,ni2023recovering}. The design of the U-shaped framework can fundamentally address this issue by placing the deeper layers of LLM at the edge devices/servers. The other direction involves the implementation of differential privacy techniques, which add controlled noise to model gradients or updates~\cite{wei2020federated}, providing theoretical privacy guarantees for SplitLoRA.

\section{Conclusion}\label{conclu}

In this paper, we propose the first SL LLM fine-tuning framework, named SplitLoRA. SplitLoRA is built on the split federated learning (SFL) framework, amalgamating the advantages of parallel training from FL and model splitting from SL and thus greatly enhancing the training efficiency. It is worth noting that SplitLoRA is the inaugural open-source benchmark for SL LLM fine-tuning, providing a foundation for research efforts dedicated to advancing SL LLM fine-tuning. Extensive simulations validate that SplitLoRA achieves target accuracy in significantly less time than centralized LLM fine-tuning frameworks, demonstrating the superior training performance of SplitLoRA. We have discussed emerging challenges and research directions in SplitLoRA, where we advocate more future efforts in this realm.

% \appendices

% % you can choose not to have a title for an appendix
% % if you want by leaving the argument blank
% \section{}

% % use section* for acknowledgment
% \section*{Acknowledgment}


\ifCLASSOPTIONcaptionsoff
  \newpage
\fi



%\bibliographystyle{IEEEtran}


% \end{thebibliography}

\bibliographystyle{IEEEtran}
\bibliography{reference}

\begin{thebibliography}{10}
\providecommand{\url}[1]{#1}
\csname url@samestyle\endcsname
\providecommand{\newblock}{\relax}
\providecommand{\bibinfo}[2]{#2}
\providecommand{\BIBentrySTDinterwordspacing}{\spaceskip=0pt\relax}
\providecommand{\BIBentryALTinterwordstretchfactor}{4}
\providecommand{\BIBentryALTinterwordspacing}{\spaceskip=\fontdimen2\font plus
\BIBentryALTinterwordstretchfactor\fontdimen3\font minus \fontdimen4\font\relax}
\providecommand{\BIBforeignlanguage}[2]{{%
\expandafter\ifx\csname l@#1\endcsname\relax
\typeout{** WARNING: IEEEtran.bst: No hyphenation pattern has been}%
\typeout{** loaded for the language `#1'. Using the pattern for}%
\typeout{** the default language instead.}%
\else
\language=\csname l@#1\endcsname
\fi
#2}}
\providecommand{\BIBdecl}{\relax}
\BIBdecl

\bibitem{openai2023gpt4}
{OpenAI}, ``{GPT-4 Technical Report},'' \emph{arXiv preprint arXiv:2303.08774}, 2023.

\bibitem{llama2}
H.~Touvron, L.~Martin, K.~Stone, P.~Albert, A.~Almahairi, Y.~Babaei, N.~Bashlykov, S.~Batra, P.~Bhargava, S.~Bhosale \emph{et~al.}, ``{Llama 2: Open Foundation and Fine-tuned Chat Models},'' \emph{arXiv preprint arXiv:2307.09288}, 2023.

\bibitem{palm}
A.~Chowdhery, S.~Narang, J.~Devlin, M.~Bosma, G.~Mishra, A.~Roberts, P.~Barham, H.~W. Chung, C.~Sutton, S.~Gehrmann \emph{et~al.}, ``{Palm: Scaling Language Modeling with Pathways},'' \emph{arXiv preprint arXiv:2204.02311}, 2022.

\bibitem{cardenas2024autohealth}
L.~Cardenas, K.~Parajes, M.~Zhu, and S.~Zhai, ``{AutoHealth: Advanced LLM-Empowered Wearable Personalized Medical Butler for Parkinson’s Disease Management},'' in \emph{Proc. CCWC}, 2024.

\bibitem{lin2023pushing}
Z.~Lin, G.~Qu, Q.~Chen, X.~Chen, Z.~Chen, and K.~Huang, ``{Pushing Large Language Models to the 6G Edge: Vision, Challenges, and Opportunities},'' \emph{arXiv preprint arXiv:2309.16739}, 2023.

\bibitem{berrios2023towards}
W.~Berrios, G.~Mittal, T.~Thrush, D.~Kiela, and A.~Singh, ``{Towards Language Models that Can See: Computer Vision Through the Lens of Natural Language},'' \emph{arXiv preprint arXiv:2306.16410}, 2023.

\bibitem{qiu2024ifvit}
Y.~Qiu, H.~Chen, X.~Dong, Z.~Lin, I.~Y. Liao, M.~Tistarelli, and Z.~Jin, ``{Ifvit: Interpretable Fixed-length Representation for Fingerprint Matching via Vision Transformer},'' \emph{arXiv preprint arXiv:2404.08237}, 2024.

\bibitem{wang2024visionllm}
W.~Wang, Z.~Chen, X.~Chen, J.~Wu, X.~Zhu, G.~Zeng, P.~Luo, T.~Lu, J.~Zhou, Y.~Qiao \emph{et~al.}, ``{Visionllm: Large Language Model is Also An Open-ended Decoder For Vision-centric Tasks},'' \emph{Proc. NIPS}, 2024.

\bibitem{hu2024agentscodriver}
S.~Hu, Z.~Fang, Z.~Fang, X.~Chen, and Y.~Fang, ``{AgentsCoDriver: Large Language Model Empowered Collaborative Driving with Lifelong Learning},'' \emph{arXiv preprint arXiv:2404.06345}, 2024.

\bibitem{zhou2023vision}
X.~Zhou, M.~Liu, B.~L. Zagar, E.~Yurtsever, and A.~C. Knoll, ``{Vision Language Models in Autonomous Driving and Intelligent Transportation Systems},'' \emph{arXiv preprint arXiv:2310.14414}, 2023.

\bibitem{tian2023vistagpt}
Y.~Tian, X.~Li, H.~Zhang, C.~Zhao, B.~Li, X.~Wang, and F.-Y. Wang, ``{VistaGPT: Generative Parallel Transformers for Vehicles with Intelligent Systems for Transport Automation},'' \emph{IEEE Trans. Intell. Veh.}, 2023.

\bibitem{qu2024trimcaching1}
G.~Qu, Z.~Lin, Q.~Chen, J.~Li, F.~Liu, X.~Chen, and K.~Huang, ``{TrimCaching: Parameter-sharing Edge Caching for AI Model Downloading},'' \emph{arXiv preprint arXiv:2404.14204}, 2024.

\bibitem{villalobos2022will}
P.~Villalobos, J.~Sevilla, L.~Heim, T.~Besiroglu, M.~Hobbhahn, and A.~Ho, ``{Will We Run Out of Data? An Analysis of the Limits of Scaling Datasets in Machine Learning},'' \emph{arXiv preprint arXiv:2211.04325}, 2022.

\bibitem{wang2023far}
Y.~Wang, H.~Ivison, P.~Dasigi, J.~Hessel, T.~Khot, K.~R. Chandu, D.~Wadden, K.~MacMillan, N.~A. Smith, I.~Beltagy \emph{et~al.}, ``{How Far Can Camels Go? Exploring the State of Instruction Tuning on Open Resources},'' \emph{arXiv preprint arXiv:2306.04751}, 2023.

\bibitem{xu2023wizardlm}
C.~Xu, Q.~Sun, K.~Zheng, X.~Geng, P.~Zhao, J.~Feng, C.~Tao, and D.~Jiang, ``Wizardlm: Empowering large language models to follow complex instructions,'' \emph{arXiv preprint arXiv:2304.12244}, 2023.

\bibitem{kaplan2020scaling}
J.~Kaplan, S.~McCandlish, T.~Henighan, T.~B. Brown, B.~Chess, R.~Child, S.~Gray, A.~Radford, J.~Wu, and D.~Amodei, ``{Scaling Laws for Neural Language Models},'' \emph{arXiv preprint arXiv:2001.08361}, 2020.

\bibitem{llm_medicine}
A.~J. Thirunavukarasu, D.~S.~J. Ting, K.~Elangovan, L.~Gutierrez, T.~F. Tan, and D.~S.~W. Ting, ``{Large Language Models in Medicine},'' \emph{Nature medicine}, vol.~29, no.~8, pp. 1930--1940, 2023.

\bibitem{wu2023bloomberggpt}
S.~Wu, O.~Irsoy, S.~Lu, V.~Dabravolski, M.~Dredze, S.~Gehrmann, P.~Kambadur, D.~Rosenberg, and G.~Mann, ``{Bloomberggpt: A Large Language Model for Finance},'' \emph{arXiv preprint arXiv:2303.17564}, 2023.

\bibitem{singhal2023towards}
K.~Singhal, T.~Tu, J.~Gottweis, R.~Sayres, E.~Wulczyn, L.~Hou, K.~Clark, S.~Pfohl, H.~Cole-Lewis, D.~Neal \emph{et~al.}, ``{Towards Expert-level Medical Question Answering with Large Language Models},'' \emph{arXiv preprint arXiv:2305.09617}, 2023.

\bibitem{ye2024openfedllm}
R.~Ye, W.~Wang, J.~Chai, D.~Li, Z.~Li, Y.~Xu, Y.~Du, Y.~Wang, and S.~Chen, ``{OpenFedLLM: Training Large Language Models on Decentralized Private Data via Federated Learning},'' \emph{arXiv preprint arXiv:2402.06954}, 2024.

\bibitem{cai2023efficient}
D.~Cai, Y.~Wu, S.~Wang, F.~X. Lin, and M.~Xu, ``{Efficient Federated Learning for Modern NLP},'' in \emph{Proc. Mobicom}, 2023.

\bibitem{fang2024automated}
Z.~Fang, Z.~Lin, Z.~Chen, X.~Chen, Y.~Gao, and Y.~Fang, ``{Automated Federated Pipeline for Parameter-efficient Fine-tuning of Large Language Models},'' \emph{arXiv preprint arXiv:2404.06448}, 2024.

\bibitem{lin2023fedsn}
Z.~Lin, Z.~Chen, Z.~Fang, X.~Chen, X.~Wang, and Y.~Gao, ``{FedSN: A General Federated Learning Framework over LEO Satellite Networks},'' \emph{arXiv preprint arXiv:2311.01483}, 2023.

\bibitem{konevcny2016federated}
J.~Kone{\v{c}}n{\`y}, H.~B. McMahan, F.~X. Yu, P.~Richt{\'a}rik, A.~T. Suresh, and D.~Bacon, ``{Federated Learning: Strategies for Improving Communication Efficiency},'' \emph{arXiv preprint arXiv:1610.05492}, 2016.

\bibitem{zhang2024fedac}
Y.~Zhang, H.~Chen, Z.~Lin, Z.~Chen, and J.~Zhao, ``{FedAC: A Adaptive Clustered Federated Learning Framework for Heterogeneous Data},'' \emph{arXiv preprint arXiv:2403.16460}, 2024.

\bibitem{vepakomma2018split}
P.~Vepakomma, O.~Gupta, T.~Swedish, and R.~Raskar, ``{Split Learning for Health: Distributed Deep Learning without Sharing Raw Patient Data},'' \emph{arXiv preprint arXiv:1812.00564}, 2018.

\bibitem{lin2023efficient}
Z.~Lin, G.~Zhu, Y.~Deng, X.~Chen, Y.~Gao, K.~Huang, and Y.~Fang, ``{Efficient Parallel Split Learning over Resource-constrained Wireless Edge Networks},'' \emph{{IEEE} Trans. Mobile Comput.}, 2024.

\bibitem{lyu2023optimal}
S.~Lyu, Z.~Lin, G.~Qu, X.~Chen, X.~Huang, and P.~Li, ``{Optimal Resource Allocation for U-shaped Parallel Split Learning},'' \emph{arXiv preprint arXiv:2308.08896}, 2023.

\bibitem{lin2023split}
Z.~Lin, G.~Qu, X.~Chen, and K.~Huang, ``{Split Learning in 6G Edge Networks},'' \emph{{IEEE} Wireless Commun.}, 2024.

\bibitem{thapa2022splitfed}
C.~Thapa, P.~C.~M. Arachchige, S.~Camtepe, and L.~Sun, ``{Splitfed: When Federated Learning Meets Split Learning},'' in \emph{Proc. AAAI}, 2022.

\bibitem{hu2021lora}
E.~J. Hu, Y.~Shen, P.~Wallis, Z.~Allen-Zhu, Y.~Li, S.~Wang, L.~Wang, and W.~Chen, ``{Lora: Low-rank Adaptation of Large Language Models},'' \emph{arXiv preprint arXiv:2106.09685}, 2021.

\bibitem{sheng2023s}
Y.~Sheng, S.~Cao, D.~Li, C.~Hooper, N.~Lee, S.~Yang, C.~Chou, B.~Zhu, L.~Zheng, K.~Keutzer \emph{et~al.}, ``{S-LoRA: Serving Thousands of Concurrent LoRA Adapters},'' \emph{arXiv preprint arXiv:2311.03285}, 2023.

\bibitem{radford2019language}
A.~Radford, J.~Wu, R.~Child, D.~Luan, D.~Amodei, I.~Sutskever \emph{et~al.}, ``{Language Models are Unsupervised Multitask Learners},'' \emph{OpenAI blog}, vol.~1, no.~8, p.~9, 2019.

\bibitem{novikova2017e2e}
J.~Novikova, O.~Du{\v{s}}ek, and V.~Rieser, ``{The E2E Dataset: New Challenges For End-to-End Generation},'' in \emph{Proc. SIGDIAL}, 2017.

\bibitem{dale2021gpt}
R.~Dale, ``{GPT-3: What’s It Good For?}'' \emph{Nat. Lang. Eng.}, vol.~27, no.~1, pp. 113--118, 2021.

\bibitem{sanderson2023gpt}
K.~Sanderson, ``{GPT-4 Is Here: What Scientists Think},'' \emph{Nature}, vol. 615, no. 7954, p. 773, 2023.

\bibitem{devlin2018bert}
J.~Devlin, M.-W. Chang, K.~Lee, and K.~Toutanova, ``{Bert: Pre-training of Deep Bidirectional Transformers for Language Understanding},'' \emph{arXiv preprint arXiv:1810.04805}, 2018.

\bibitem{jiang2023fdapt}
L.~Jiang, F.~Svoboda, and N.~D. Lane, ``{FDAPT: Federated Domain-adaptive Pre-training for Language Models},'' \emph{arXiv preprint arXiv:2307.06933}, 2023.

\bibitem{englhardt2023classification}
Z.~Englhardt, C.~Ma, M.~E. Morris, X.~Xu, C.-C. Chang, L.~Qin, X.~Liu, S.~Patel, V.~Iyer \emph{et~al.}, ``{From Classification to Clinical Insights: Towards Analyzing and Reasoning About Mobile and Behavioral Health Data With Large Language Models},'' \emph{arXiv preprint arXiv:2311.13063}, 2023.

\bibitem{nan2021deep}
Z.~Nan, H.~Guan, X.~Shen, and C.~Liao, ``{Deep NLP-Based Co-evolvement for Synthesizing Code Analysis from Natural Language},'' in \emph{Proc. of CC}, 2021.

\bibitem{wang2023drivemlm}
W.~Wang, J.~Xie, C.~Hu, H.~Zou, J.~Fan, W.~Tong, Y.~Wen, S.~Wu, H.~Deng, Z.~Li \emph{et~al.}, ``{DriveMLM: Aligning Multi-modal Large Language Models with Behavioral Planning States for Autonomous Driving},'' \emph{arXiv preprint arXiv:2312.09245}, 2023.

\bibitem{cui2024survey}
C.~Cui, Y.~Ma, X.~Cao, W.~Ye, Y.~Zhou, K.~Liang, J.~Chen, J.~Lu, Z.~Yang, K.-D. Liao \emph{et~al.}, ``{A Survey on Multimodal Large Language Models for Autonomous Driving},'' in \emph{Proc. WCAC}, 2024.

\bibitem{mai2023llm}
J.~Mai, J.~Chen, B.~Li, G.~Qian, M.~Elhoseiny, and B.~Ghanem, ``{LLM As A Robotic Brain: Unifying Egocentric Memory and Control},'' \emph{arXiv preprint arXiv:2304.09349}, 2023.

\bibitem{kannan2023smart}
S.~S. Kannan, V.~L. Venkatesh, and B.-C. Min, ``{Smart-LLM: Smart Multi-agent Robot Task Planning Using Large Language Models},'' \emph{arXiv preprint arXiv:2309.10062}, 2023.

\bibitem{hu2023adaptive}
S.~Hu, Z.~Fang, H.~An, G.~Xu, Y.~Zhou, X.~Chen, and Y.~Fang, ``{Adaptive Communications in Collaborative Perception with Domain Alignment for Autonomous Driving},'' \emph{arXiv preprint arXiv:2310.00013}, 2023.

\bibitem{lin2022channel}
Z.~Lin, L.~Wang, J.~Ding, B.~Tan, and S.~Jin, ``Channel power gain estimation for terahertz vehicle-to-infrastructure networks,'' \emph{{IEEE} Commun. Lett.}, vol.~27, no.~1, pp. 155--159, 2022.

\bibitem{hu2024collaborative}
S.~Hu, Z.~Fang, Y.~Deng, X.~Chen, and Y.~Fang, ``{Collaborative Perception for Connected and Autonomous Driving: Challenges, Possible Solutions and Opportunities},'' \emph{arXiv preprint arXiv:2401.01544}, 2024.

\bibitem{mcmahan2017communication}
B.~McMahan, E.~Moore, D.~Ramage, S.~Hampson, and B.~A. y~Arcas, ``{Communication-efficient Learning of Deep Networks From Decentralized Data},'' in \emph{Proc. AISTATS}, 2017.

\bibitem{lin2024adaptsfl}
Z.~Lin, G.~Qu, W.~Wei, X.~Chen, and K.~K. Leung, ``{AdaptSFL: Adaptive Split Federated Learning in Resource-constrained Edge Networks},'' \emph{arXiv preprint arXiv:2403.13101}, 2024.

\bibitem{wu2023split}
W.~Wu, M.~Li, K.~Qu, C.~Zhou, X.~Shen, W.~Zhuang, X.~Li, and W.~Shi, ``{Split Learning over Wireless Networks: Parallel Design and Resource Management},'' \emph{{IEEE} J. Select. Areas Commun.}, vol.~41, no.~4, pp. 1051--1066, 2023.

\bibitem{liu2022wireless}
X.~Liu, Y.~Deng, and T.~Mahmoodi, ``{Wireless Distributed Learning: A New Hybrid Split and Federated Learning Approach},'' \emph{{IEEE} Trans. Wireless Commun.}, vol.~22, no.~4, pp. 2650--2665, 2022.

\bibitem{cheng2023cheese}
Z.~Cheng, X.~Xia, M.~Liwang, X.~Fan, Y.~Sun, X.~Wang, and L.~Huang, ``{CHEESE: Distributed Clustering-based Hybrid Federated Split Learning over Edge Networks},'' \emph{{IEEE} Trans. Parallel Distrib. Syst.}, 2023.

\bibitem{huawei2019}
{Huawei}, \emph{{NET4AI: Supporting AI as a Service in 6G}}.\hskip 1em plus 0.5em minus 0.4em\relax Cambridge, U.K.: Cambridge Univ. Press, 2022.

\bibitem{he2016deep}
K.~He, X.~Zhang, S.~Ren, and J.~Sun, ``{Deep Residual Learning for Image Recognition},'' in \emph{Proc. CVPR}, 2016, pp. 770--778.

\bibitem{simonyan2014very}
K.~Simonyan and A.~Zisserman, ``{Very Deep Convolutional Networks for Large-scale Image Recognition},'' in \emph{Proc. ICLR}, 2015.

\bibitem{houlsby2019parameter}
N.~Houlsby, A.~Giurgiu, S.~Jastrzebski, B.~Morrone, Q.~De~Laroussilhe, A.~Gesmundo, M.~Attariyan, and S.~Gelly, ``{Parameter-Efficient Transfer Learning for {NLP}},'' in \emph{Proc. ICML}, 2019, pp. 2790--2799.

\bibitem{pfeiffer2020adapterfusion}
J.~Pfeiffer, A.~Kamath, A.~R{\"u}ckl{\'e}, K.~Cho, and I.~Gurevych, ``{AdapterFusion: Non-Destructive Task Composition for Transfer Learning},'' in \emph{Proc. EACL}, 2021, pp. 487--503.

\bibitem{karimi2021compacter}
R.~Karimi~Mahabadi, J.~Henderson, and S.~Ruder, ``{Compacter: Efficient Low-Rank Hypercomplex Adapter Layers},'' 2021, pp. 1022--1035.

\bibitem{li2023prompt}
L.~Li, Y.~Zhang, and L.~Chen, ``{Prompt Distillation for Efficient LLM-based Recommendation},'' in \emph{Proc. CIKM}, 2023, pp. 1348--1357.

\bibitem{li2021prefix}
X.~L. Li and P.~Liang, ``{Prefix-tuning: Optimizing Continuous Prompts for Generation},'' \emph{arXiv preprint arXiv:2101.00190}, 2021.

\bibitem{liu2021p}
X.~Liu, K.~Ji, Y.~Fu, W.~L. Tam, Z.~Du, Z.~Yang, and J.~Tang, ``{P-tuning v2: Prompt Tuning Can Be Comparable to Fine-tuning Universally Across Scales and Tasks},'' \emph{arXiv preprint arXiv:2110.07602}, 2021.

\bibitem{aghajanyan2021intrinsic}
A.~Aghajanyan, S.~Gupta, and L.~Zettlemoyer, ``{Intrinsic Dimensionality Explains the Effectiveness of Language Model Fine-Tuning},'' in \emph{Proc. IJCNLP}, 2021, pp. 7319--7328.

\bibitem{fedllm-position}
C.~Chen, X.~Feng, J.~Zhou, J.~Yin, and X.~Zheng, ``{Federated Large Language Model: A Position Paper},'' \emph{arXiv preprint arXiv:2307.08925}, 2023.

\bibitem{fan2023fate}
T.~Fan, Y.~Kang, G.~Ma, W.~Chen, W.~Wei, L.~Fan, and Q.~Yang, ``{Fate-LLM: A Industrial Grade Federated Learning Framework for Large Language Models},'' \emph{arXiv preprint arXiv:2310.10049}, 2023.

\bibitem{federatedscopellm}
W.~Kuang, B.~Qian, Z.~Li, D.~Chen, D.~Gao, X.~Pan, Y.~Xie, Y.~Li, B.~Ding, and J.~Zhou, ``{FederatedScope-LLM: A Comprehensive Package for Fine-tuning Large Language Models in Federated Learning},'' \emph{arXiv preprint arXiv:2309.00363}, 2023.

\bibitem{fedit}
J.~Zhang, S.~Vahidian, M.~Kuo, C.~Li, R.~Zhang, G.~Wang, and Y.~Chen, ``{Towards Building the Federated GPT: Federated Instruction Tuning},'' \emph{arXiv preprint arXiv:2305.05644}, 2023.

\bibitem{pasquini2021unleashing}
D.~Pasquini, G.~Ateniese, and M.~Bernaschi, ``{Unleashing the Tiger: Inference Attacks on Split Learning},'' in \emph{Proc. CCS}, 2021.

\bibitem{qu2024trimcaching}
G.~Qu, Z.~Lin, F.~Liu, X.~Chen, and K.~Huang, ``{TrimCaching: Parameter-sharing AI Model Caching in Wireless Edge Networks},'' in \emph{Proc. ICDCS}, 2024.

\bibitem{wang2023can}
B.~Wang, Y.~J. Zhang, Y.~Cao, B.~Li, H.~B. McMahan, S.~Oh, Z.~Xu, and M.~Zaheer, ``{Can Public Large Language Models Help Private Cross-device Federated Learning?}'' \emph{arXiv preprint arXiv:2305.12132}, 2023.

\bibitem{xia2023sheared}
M.~Xia, T.~Gao, Z.~Zeng, and D.~Chen, ``{Sheared llama: Accelerating Language Model Pre-training Via Structured Pruning},'' \emph{arXiv preprint arXiv:2310.06694}, 2023.

\bibitem{dettmers2024qlora}
T.~Dettmers, A.~Pagnoni, A.~Holtzman, and L.~Zettlemoyer, ``{Qlora: Efficient Finetuning of Quantized LLMs},'' \emph{Proc. NIPS}, 2024.

\bibitem{nasr2019comprehensive}
M.~Nasr, R.~Shokri, and A.~Houmansadr, ``{Comprehensive Privacy Analysis of Deep Learning: Passive and Active White-box Inference Attacks Against Centralized and Federated Learning},'' in \emph{Proc. SP}, 2019.

\bibitem{gupta2022recovering}
S.~Gupta, Y.~Huang, Z.~Zhong, T.~Gao, K.~Li, and D.~Chen, ``{Recovering Private Text in Federated Learning of Language Models},'' \emph{Proc. NIPS}, 2022.

\bibitem{carlini2021extracting}
N.~Carlini, F.~Tramer, E.~Wallace, M.~Jagielski, A.~Herbert-Voss, K.~Lee, A.~Roberts, T.~Brown, D.~Song, U.~Erlingsson \emph{et~al.}, ``{Extracting Training Data From Large Language Models},'' in \emph{USENIX Security}, 2021, pp. 2633--2650.

\bibitem{ni2023eavesdropping}
T.~Ni, G.~Lan, J.~Wang, Q.~Zhao, and W.~Xu, ``{Eavesdropping Mobile App Activity via Radio-Frequency Energy Harvesting},'' in \emph{Proc. USENIX Security Symposium}, 2023.

\bibitem{ni2023recovering}
T.~Ni, X.~Zhang, and Q.~Zhao, ``{Recovering Fingerprints from In-Display Fingerprint Sensors via Electromagnetic Side Channel},'' in \emph{Proc. ACM CCS}, 2023.

\bibitem{wei2020federated}
K.~Wei, J.~Li, M.~Ding, C.~Ma, H.~H. Yang, F.~Farokhi, S.~Jin, T.~Q. Quek, and H.~V. Poor, ``{Federated Learning with Differential Privacy: Algorithms and Performance Analysis},'' \emph{IEEE Trans. Inf. Forensics Secur.}, vol.~15, pp. 3454--3469, 2020.

\end{thebibliography}


% \end{IEEEbiography}


% % insert where needed to balance the two columns on the last page with
% %\newpage



%\vfill


\end{document}